%% file: main.tex
\pdfoutput=1

\documentclass[11pt]{article}

\usepackage[final]{acl}

\usepackage{times}
\usepackage{latexsym}
\usepackage{xspace}

\usepackage[T1]{fontenc}

\usepackage[utf8]{inputenc}

\usepackage{microtype}

\usepackage{inconsolata}

\usepackage{graphicx}
\usepackage{booktabs}
\usepackage{multirow}
\usepackage{colortbl}
\usepackage[most]{tcolorbox}

\title{Operationalizing AI for Good: Spotlight on Deployment and Integration of AI Models in Humanitarian Work}

\input{dfn}

\author{
  Anton Abilov, Ke Zhang, Hemank Lamba, \\
  \textbf{Elizabeth M. Olson, Joel Tetreault, Alex Jaimes} \\
  Dataminr, Inc. \\
  \texttt{\{aabilov,kzhang,hlamba,elizabeth.olson,jtetreault,ajaimes\}} \\\texttt{@dataminr.com}
}

\begin{document}
\maketitle

\begin{abstract}

Publications in the AI for Good space have tended to focus on the research and model development that can support high-impact applications.  However, very few AI for Good papers discuss the process of deploying and collaborating with the partner organization, 
and the resulting real-world impact.  
In this work, we share details about the close collaboration with a humanitarian-to-humanitarian (H2H) organization and how to not only deploy the AI model in a resource-constrained environment, but also how to maintain it for continuous performance updates, and share key takeaways for practitioners.

\end{abstract}

\input{00intro}

\input{problem_overview}

\input{10implementation}
\input{deployment}

\input{31discussion}

\section{Limitations}
We acknowledge that this is just one example of an AI deployment in a humanitarian setting.  Ideally, we would present several examples of such deployments to paint a more robust picture of the different decisions partners can make, and the associated challenges.  However, that is outside the scope of this short paper.  We hope that by going into the details of this deployment process and showing the real-world impact will encourage others to publish their findings as well.

Another aspect we want to acknowledge is that there are many different types of AI for Good projects and deployments.  A group of AI scientists partnering with a humanitarian organization is just one configuration.

\section{Ethical Considerations}
The dataset is constructed from publicly available news articles, ensuring that no contractual agreements were violated in the data acquisition process. Our web scraper strictly accessed openly available content, excluding any material behind paywalls.
For the annotation process, we engaged internal humanitarian experts from the partnering organization. These experts were fairly compensated as part of their professional, paid employment.

\bibliography{custom}

\appendix
\input{appendix}

\end{document}

%% file: dfn.tex
\newcommand{\insecurityinsight}{Insecurity Insight\xspace}

%% file: 00intro.tex
\section{Introduction}

The last ten years have seen a surge in AI and Natural Language Processing research to address real world problems that have a social good impact \cite{adauto-etal-2023-beyond}.  Many of these problems align with the United Nations Sustainable Development Goals (UNSDG)\footnote{https://sdgs.un.org/goals}.  
This has also led to a surge in publications in this space to the point that
even prominent AI research conferences have special tracks and themes related to social good (ie. AAAI, ACL-IJCNLP in 2021 \cite{acl-2021-association}) and many targeted venues to tackle this topic such as the NLP for Positive Impact workshop series\footnote{https://sites.google.com/view/nlp4positiveimpact}.

\citet{jin-etal-2021-good} describe four different stages of AI for Good tasks: 1. Fundamental theories, 2. Building block tools, 3. Applicable tools and 4. Deployed applications.  While there have been a lot of publications in this space (for example \citet{adauto-etal-2023-beyond} found that just over 13\% of all papers in the ACL Anthology map to one of the UNSDGs), most published AI for Good work has tended to focus more on the first three stages: specifically on analysis of the problem area, building a dataset, or building a model.  However, there is comparatively very little published work on the fourth stage: on how these models fare when deployed in the real world and how they align with the expectations of the social good organization.  In fact, for the ACL-IJCNLP 2021 special theme of "NLP for Social Good", only one of the twelve accepted papers mentioned deployment.

In addition, there has been very little work that discusses the collaboration process between a humanitarian organization and AI practitioners where a model is built to be used by the partner organization. The closest works are~\citet{tomavsev2020ai} and~\citet{10.1145/3461702.3462599}, which highlight how AI teams should approach and undertake AI4SG projects - but do not mention any details about development and deployment process.

In this short paper, we present our experience with working with  \insecurityinsight\footnote{\href{https://insecurityinsight.org/}{https://insecurityinsight.org/}}, a humanitarian-to-humanitarian organization (H2H), to bring an NLP model into the real world and provide impact to that organization and the aid community it supports. 
This work builds upon our previous research~\cite{lamba-etal-2024-humvi}, in which we developed a multilingual dataset of news articles in English, French, and Arabic, annotated with various types of violent incidents categorized by the humanitarian sectors they affect—such as aid security, education, food security, health, and protection. We also evaluated a range of deep learning architectures and techniques to tackle the associated task-specific challenges. 
In this paper, we take the next step by addressing the critical final stage: model deployment. 
In particular, we discuss not only the technical and process aspects of deploying a model in a resource-constrained environment, but also how to maintain it for continuous performance updates.  We conclude with key takeaways and best practices for both AI model developers and humanitarian experts around technical topics, collaboration and processes.  While this is just one example of a deployment, we hope this paper will encourage others to share their experiences and lessons learned.

%% file: problem_overview.tex
\section{Partnership Case Study}

\subsection{Partner Details}
\insecurityinsight is a data-based H2H organization. Their aim is to support the work of aid agencies, healthcare providers, and other civil organizations by providing data-driven intelligence reports that can be used by these organizations for efficient resource allocation, humanitarian response, fund raising, advocacy, among others. Before our collaboration, \insecurityinsight collected news articles from select data sources (i.e. NewsAPI~\cite{lisivick2018newsapi}, OSAC\footnote{\href{https://www.osac.gov/Content/Browse/News}{https://www.osac.gov/Content/Browse/News}}, and through manual uploading of news articles by humanitarian experts. These articles were then passed to an SVM model for relevance classification and category classification (categories defined on downstream humanitarian impact - education, aid, health and protection). Once classified and tagged, they were reviewed and summarized by humanitarian experts.  
However, this workflow had two drawbacks: (1) it was limited to existing downstream humanitarian categories and (2) it focused only on English articles.

\subsection{Problem Scope}
For our partnership with \insecurityinsight, we identified the following three shared goals.  The plan was to develop NLP models which could address these goals and then deploy them in their workflow.

\noindent
\textbf{Goal 1.} Improve the existing workflow to identify and classify more relevant news events.

\noindent
\textbf{Goal 2.} Expand to new domain of food security.

\noindent
\textbf{Goal 3.} Expand to French and Arabic articles.

\subsection{Resource Constraints}
A key challenge of AI4SG collaborations is that often the organization that uses AI might not have many resources to dedicate to the development, hosting, and maintenance of AI models. Our partner organization also faced similar challenges. Working in resource-constrained environments produces interesting challenges for AI developers. We list some of them below:

\noindent
\textbf{Labeling Resources}: Our partner had a limited number of humanitarian experts on staff, leading to a constrained article review capacity in the live production workflow, as well as limited time for completing separate offline annotation tasks, which were crucial for model development. 

\noindent
\textbf{Low Compute Environment}: The model was intended to be deployed within the existing infrastructure to avoid incurring additional costs for the partner organization. 
The deployment infrastructure consists of Heroku Basic dyno (1 vCPU, 512MB memory) for running scheduled crawling jobs, a dedicated VPS machine (4 vCPUs, 8GB memory) for hosting the classifier API and a MongoDB database (2GB storage).
There is no real-time latency requirement for the model inference, however it is critical for the throughput rate of the scheduled crawling and classification jobs to keep up with the influx of new articles.

\noindent
\textbf{Maintenance}: The partner had minimal engineering staff so it was crucial to deliver a solution that was robust and easily maintained.

%% file: deployment.tex
\section{Implementation and Deployment}

\begin{figure*}[!hbtp]
\centering
    \includegraphics[width=0.9\textwidth]{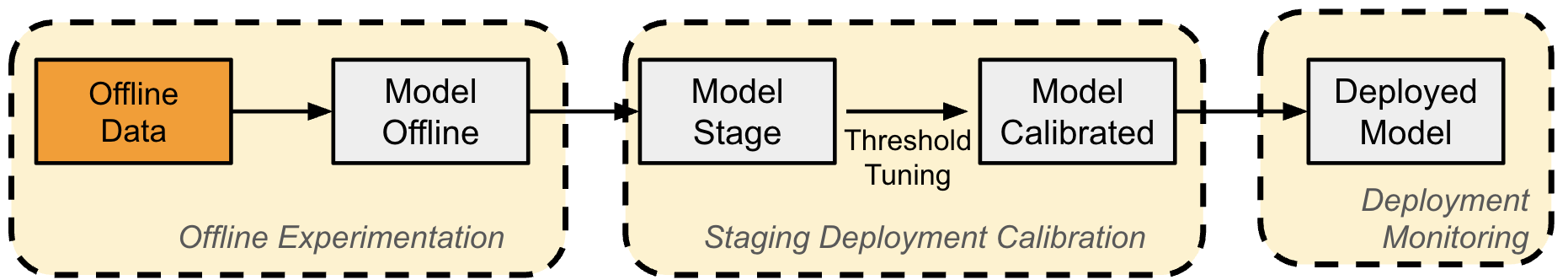}
    \caption{Stages of our model lifecycle}
    \label{fig:lifecycle}
\end{figure*}

Following standard ML Ops practices \cite{Shankar_2024} we split the model development into three stages: offline experimentation, staging 
deployment calibration and deployment monitoring (as presented in Figure \ref{fig:lifecycle}).

\subsection{Offline Experimentation}

{\bf GDELT Source Expansion}: Two of the key goals are to expand the current workflow so that it can tag in new domains and expand to articles in French and Arabic. 
To address both, we
augment the current data sourcing with GDELT~\cite{leetaru2013gdelt}, a large
real-time open-source database of multilingual news articles. 

\noindent
\textbf{Data Labeling}: 
To collect labeled data for the new input distribution, we established an offline spreadsheet labeling process with 7 humanitarian experts from \insecurityinsight using annotation guidelines similar to their established live workflow.
Expert annotators reviewed the title and content of the scraped article before determining whether the article is relevant and assigning the event categories. To ensure high quality labels, we used annotator deliberation to improve high inter-annotator agreement rates. Given the limited annotation resources, we tried to get annotation for a sample of data ensuring that it was diverse in language, categories, and a base model confidence's score. The dataset and associated repository are published at 
\href{https://github.com/dataminr-ai/humvi-dataset}{https://github.com/dataminr-ai/humvi-dataset}. More details on the data collection and quality control can be found in our previous work \cite{lamba-etal-2024-humvi}.

\noindent
{\bf Model Development and Selection}: 
We trained two models - (1) Relevance Model for identifying relevant news articles, and (2) Categorization Model for tagging relevant articles with proper downstream humanitarian categories.
In order to detect food security events, the category classification model is expanded to five output classes. During training, we translated English data to French and Arabic to augment initial training samples, and used label loss masking~\cite{duarte2021plm} to account for the  new category label.
We focused on evaluating three smaller-sized multilingual transformer models - BERT~\cite{devlin2018bert}, RoBERTa~\cite{liu2019roberta, conneau2019unsupervised}, and DistilBERT~\cite{sanh2019distilbert}, all of which could be deployed given the compute and latency constraints. We temporally split the labeled data to establish offline relevance and category classification performance on a held-out test set. XLM-RoBERTa performed best in expanding to the new input domain and languages (Relevance F1 scores ranged from 0.81 to 0.83 for the three languages)  
and thus was selected for deployment ($\mathcal{M}_{stage}$); ensuring that new workflow can source new types of articles with higher coverage and can tag them for new languages and new categories.

\subsection{Staging Deployment Calibration}
Though the deployed model performed well on the offline dataset, the main test was whether those scores would hold when deployed in the real world setting and bring value to \insecurityinsight.  
We envisioned the model performance could be lower due to (1) content drift~\cite{5975223} given the offline test set was collected a few months earlier; (2) possible mismatch between offline and online computing environments; and (3) the increased volume of articles could overwhelm the human review system given limited staffing.

\noindent
{\bf Offline Test Setup}: To minimize the risks above, we worked closely with our partner to conduct a pre-deployment test in a staging environment. We integrated the GDELT data source and deployed the model $\mathcal{M}_{stage}$ and ran it in parallel to the existing production system for $2$ weeks. 
To evaluate the ``live'' performance on data from GDELT, we sampled $1,000$ examples using stratified sampling by discretized model confidence scores. 
For existing sources NewsAPI and OSAC, we re-use the labels from the production SVM-based system.  

\noindent
\textbf{Model Threshold Tuning}: 
We tuned relevance classification thresholds for each language given the annotated data sampled from the live staging environment. 
Table~\ref{Table:english_threshold_proposal} presents the recall, precision, and estimated volume of weekly articles to review given different threshold options for English. 
Table \ref{Table:english_threshold_proposal_source_breakdown} further presents the estimated volume of articles to review (i.e., articles predicted as relevant) across three different sources: NewsAPI, OSAC and GDELT. After the source expansion, around 90\% of the ingested data came from GDELT.

\begin{table}[!hbtp]
\centering
\small{
\begin{tabular}{c|c|cc|c}
Option & Threshold & Recall & Precision & Volume \\
\midrule
Baseline & 0.184 & 0.85 & 0.785 & 951 (\textbf{20x}) \\
\midrule
Option 1 & 0.646 & 0.790 & 0.802 & 803 (\textbf{17x})  \\
Option 2 & 0.943 & 0.532 & 0.854 & 484 (\textbf{10x})   \\
\bf{*}Option 3 & 0.951 & 0.405 & 0.903 & 367 (\textbf{8x})   \\
\end{tabular}
}
\caption{Volume (number of articles to review per week) and quality (precision \& recall) impact given different proposed thresholds for relevance classification for English. *=Model Selected}
\label{Table:english_threshold_proposal}
\end{table}

Per the initial requirement from our partner, the baseline model threshold ($0.184$) was tuned with max precision at minimum recall $0.85$ to minimize missing potentially relevant articles. 
With the inclusion of GDELT this approach would lead to a \textbf{20x} estimated increase (from $46$ to $951$ weekly) in articles to review.
We discussed this recall-volume trade-off with \insecurityinsight
and decided to move forward with Option 3 (henceforth $\mathcal{M}_{prod}$) at minimum $0.90$ precision to reduce the expected labeling burden increase to \textbf{8x}.
We perform a similar analysis for Arabic and French (see results in Tables \ref{Table:french_threshold_proposal} and~\ref{Table:arabic_threshold_proposal} in Appendix~\ref{sec:appendix_threshold_tuning}), and select a threshold at a lower minimum precision ($0.80$ for Arabic and $0.62$ for French) due to the smaller number of articles crawled.

\begin{table}[h]
\centering
\small{
\begin{tabular}{c|cc|cc}
Threshold & Recall & Precision & Source & Volume \\
\midrule
\multirow{3}{*}{0.184} & \multirow{3}{*}{0.85} & \multirow{3}{*}{0.785} & NewsAPI & 80 \\
& & & OSAC & 21 \\
& & & GDELT & 850 \\
\midrule
\multirow{3}{*}{0.646} & \multirow{3}{*}{0.790} & \multirow{3}{*}{0.802} & NewsAPI & 67 \\
& & & OSAC & 16 \\
& & & GDELT & 720 \\
\midrule
\multirow{3}{*}{0.943} & \multirow{3}{*}{0.532} & \multirow{3}{*}{0.854} & NewsAPI & 36 \\
& & & OSAC & 8 \\
& & & GDELT & 440 \\
\midrule
\multirow{3}{*}{\bf{*}0.951} & \multirow{3}{*}{0.405} & \multirow{3}{*}{0.903} & NewsAPI & 22 \\
& & & OSAC & 5 \\
& & & GDELT & 340 \\
\end{tabular}
}
\caption{Volume (number of articles to review per week) and quality (precision \& recall) impact given different proposed thresholds for relevance classification for English. The volume is broken down by source. Most articles came from the expanded source GDELT.  *=Model Selected}
\label{Table:english_threshold_proposal_source_breakdown}
\end{table}

For category classification, we set one threshold across all languages for each category. 
We tune it to minimum precision $>= 0.8$ in line with the baseline system.

\subsection{Post-Deployment Analysis}
To assess the deployment we compared data from the live system $4$ months after the final model $\mathcal{M}_{prod}$ deployment with the baseline system performance in 2024. 
Table~\ref{Table:post_deployment_diff} shows the impact of the deployment in terms of article volume across each stage of the system.
Overall, we surfaced \textbf{3.6x} more confirmed relevant articles compared to the baseline system with a \textbf{3.2x} increase in manual labeling effort.
The precision of the system had improved from the $0.80$ baseline and is closely aligned with the estimated precision from the pre-deployment threshold tuning stage ($0.92$ for English, $0.82$ for French and $0.82$ for Arabic).
The GDELT source expansion led to a \textbf{23x} increase in crawled articles per week, and the updated classifier predicted \textbf{9x} more articles as relevant.
A significant number of confirmed relevant articles were surfaced in French and Arabic ($42\%$ of the total baseline volume). 

\begin{table}[!hbtp]
\centering
\begin{tabular}{l|c|cc|c|}
Pipeline Stage & Baseline & Deployment \\
\midrule
Crawled & $450$ & $10,550$ \\
\midrule
Predicted Relevant & $54$ & $496$  \\
-- English & $54$ & $326$ \\
-- French & $0$ & $41$ \\
-- Arabic & $0$ & $129$ \\
\midrule
Confirmed Relevant & $43/54$ & $154/171$ \\
-- English & $43/54$ & $131/142$ \\
-- French & $0$ & $9/11$ \\
-- Arabic & $0$ & $14/17$ \\
\end{tabular}
\caption{Volume (number of articles per week) across each stage of the system before and after the model deployment.}
\label{Table:post_deployment_diff}
\end{table}

\noindent
{\bf Food Security}: We expected an \textbf{8x} volume increase but only marginally improved the system's ability to surface more articles of this category (from 1 to 3 per week). The F1 Score for this class significantly drops between offline evaluation ($F1=0.679$) and product deployment ($F1=0.014$) for English articles. And there were even no articles in Arabic or French labeled. Full results per category are presented in Table \ref{Table:model_comparison_before_and_after} in Appendix \ref{appendix:categorization_model_performance}. 
Upon further review, we determined that there were missing labels due to annotation inconsistencies, which were traced back to unclear annotator guidance and poor calibration.
This highlights the importance of performing regular data quality checks.

\begin{figure}[!tbp]
    \centering
    \includegraphics[width=\linewidth]{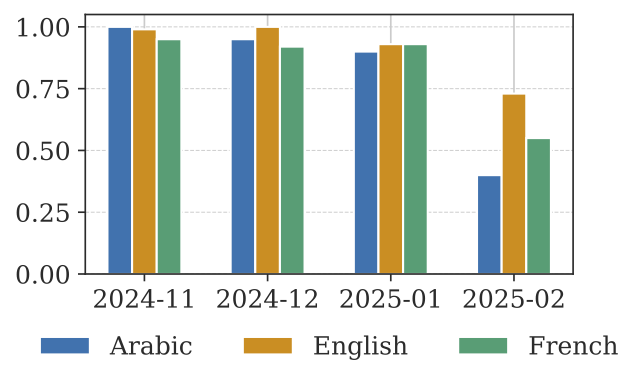}
    \caption{Relevance classifier precision over time by source language.}
    \label{fig:precision_over_time}
\end{figure}

\noindent
{\bf Performance Over Time}:
Figure \ref{fig:precision_over_time} shows the relevance model performance over time. 
Notably there was a performance drop in the last month of collected data across all languages.
This showed that there was a risk of model performance degradation due to shifts in the live data distribution.
We addressed this drop by providing the partner with workflows for continuously monitoring the live model performance and a recipe for retraining the model artifact based on new labeled data.

%% file: 31discussion.tex
\section{Discussion}
Developing and deploying AI strategies for ``AI for Good'' projects presents unique opportunities and challenges for AI practitioners and NGOs. Ensuring a sustainable and impactful deployment requires a collaborative approach that bridges technical expertise and domain-specific knowledge. 
Below we outline key takeaways from our 
collaboration with \insecurityinsight.

\noindent
\textbf{T1. Understanding the Problem}: 
Before developing AI models, practitioners must deeply understand the problem they want to address. This requires a thorough stakeholder engagement, data assessment, and problem scoping. During the early phase of the project, we gathered crucial domain knowledge from domain experts and engineers in \insecurityinsight to get a deep understanding of their current service and system, impact measurement, specific needs with priorities, resource and operational constraints, data availability and technical stack. This helps inform our key decisions in the steps of data collection, model selection and deployment. 

\noindent
{\bf T2. Data Availability and Quality}: 
Both parties must assess the availability, reliability and bias of data sources. Available data may be noisy or limited in scope, thus requiring new data collection methods or annotation. Data quality could be an ongoing issue, and thus it is important to start early and iterate: practitioners should work with domain experts to come up with clear annotation guidelines.  
In this particular study, we found it is essential to be mindful of the domain expert's time (operational cost). This requires both teams to  setup realistic and meaningful plans and schedules.

\noindent
{\bf T3. Capacity Building}: 
For AI solutions to be sustainable, partner organizations must have the capacity to use and maintain them. It is important to keep the partner in the loop throughout the development process %
and establish support mechanisms for model updates, debugging, and continuous improvement.

\noindent
{\bf T4. Model Performance Mismatch Awareness:} 
Both parties should be aware of potential discrepancies between offline evaluations and real-world AI performance (as we saw with our food security results). 
Establishing a staged testing environment helps validate and refine AI solutions before deployment, reducing unexpected behaviors in production. Both parties should be flexible in adjusting metrics  to better fit real-world needs (e.g., optimize for precision instead of recall).

\noindent
{\bf T5. Impact Assessment and Continuous Monitoring}: 
It is important to establish clear metrics to measure success. 
Once deployed, AI solutions should be regularly evaluated for performance drift (as shown in  Figure~\ref{fig:precision_over_time}). While automated monitoring pipelines can track key metrics in real-world use, continuous calibration of labeling quality is integral to informing robust metrics. Retraining with fresh data and adjusting decision thresholds helps maintain accuracy and thwart content drift.

In short, this paper details our experience of developing and deploying a model to assist a humanitarian organization in a resource-constrained setting.  
The implementation process and takeaways may be useful for practitioners that are seeking to operationalize AI models in low-resource settings. 
This  ``final stage'' is often quite challenging, and we hope other practitioners will publish their process and impacts as well.

%% file: appendix.tex
\section{Appendix}
\label{sec:appendix}

\begin{table*}[!bht]
    \begin{tabular}{|l|c|c|c|c|c|c|c|}
        \hline
        \multirow{2}{*}{Category} & \multicolumn{1}{c|}{\textbf{Old Model}} & \multicolumn{3}{c|}{\textbf{New Model (Offline Test Set)}} & \multicolumn{3}{c|}{\textbf{New Model (Live data)}} \\ 
        \cline{2-8}
        \multicolumn{1}{|c|}{} & \textbf{English} & \textbf{English} & \textbf{French} & \textbf{Arabic} & \textbf{English} & \textbf{French} & \textbf{Arabic} \\ 
        \hline
        \textbf{Food Security} & Not supported & 0.679 & 0.491 & 0.661 & \cellcolor{red!20}{0.014} & \cellcolor{orange!20}No labels & \cellcolor{orange!20}No labels \\ 
        \hline
        \textbf{Aid Security} & 0.560 & 0.729 & 0.745 & 0.688 & 0.672 & 0.947 & \cellcolor{red!10}{0.362} \\ 
        \hline
        \textbf{Education} & 0.245 & 0.773 & 0.563 & 0.571 & 0.669 & 0.671 & 0.772 \\ 
        \hline
        \textbf{Health} & 0.365 & 0.681 & 0.792 & 0.629 & 0.758 & 0.680 & 0.664 \\ 
        \hline
        \textbf{Protection} & 0.357 & 0.708 & 0.775 & 0.888 & 0.908 & 0.655 & 0.764 \\ 
        \hline
    \end{tabular}
    \caption{The performance of category classification using the offline test set versus using the live labeled data in production system. There observed as huge discrepancy of performance metrics for Food Security across the languages, and Aid Security in Arabic language.}
    \label{Table:model_comparison_before_and_after}
\end{table*}

\subsection{Threshold Tuning across 3 Languages}
\label{sec:appendix_threshold_tuning}

As we tune the thresholds per language, Table \ref{Table:french_threshold_proposal} and \ref{Table:arabic_threshold_proposal} presents the quality and volume impact under different thresholds. Arabic shows a good volume of articles, which meets well with our initial goal of expanding to collecting articles from Arabic-speaking local geographical areas. Although we were not able to surface a good number of French articles, this is still a good start for \insecurityinsight. 

\begin{table}[!bht]
\centering
\small{
\begin{tabular}{c|c|cc|c}
Option & Threshold & Recall & Precision & Volume \\
\midrule
Baseline & NA & NA & NA & 0 \\
\midrule
Option 1        & \texttt{0.125} & 0.676    & 0.50 & 63 \\
\bf{*}Option 2  & \texttt{0.881} & 0.432    & 0.615 & 39 \\
Option 3        & \texttt{0.942} & 0.324    & 0.706 & 26   \\
\end{tabular}
}
\caption{Volume (number of articles to review per week) and quality (precision \& recall) impact given different proposed thresholds for relevance classification for French, which was crawled only from GDELT source. *=Model Selected}
\label{Table:french_threshold_proposal}
\end{table}

\begin{table}[!bht]
\centering
\small{
\begin{tabular}{c|c|cc|c}
Option & Threshold & Recall & Precision & Volume \\
\midrule
Baseline & NA & NA & NA & 0 \\
\midrule
Option 1        & \texttt{0.361} & 0.793    & 0.605 & 230 \\
Option 2        & \texttt{0.824} & 0.690    & 0.714 & 211 \\
\bf{*}Option 3  & \texttt{0.952} & 0.414    & 0.8 & 150  \\
\end{tabular}
}
\caption{Volume (number of articles to review per week) and quality (precision \& recall) impact given different proposed thresholds for relevance classification for Arabic, which was crawled only from GDELT source.  *=Model Selected}
\label{Table:arabic_threshold_proposal}
\end{table}

\subsection{Categorization Model Performance}
\label{appendix:categorization_model_performance}
Table \ref{Table:model_comparison_before_and_after} compares the metrics of categorization model between using the offline test set and using the live labeled data in production. The metrics across most event category and languages align well before and after deployment. However, we observed significant metric discrepancy for Food Security across all languages, and for Aid Security in Arabic. This could be attributed to multiple reasons: (1) model degenerates due to content drifts and poor model generalization; (2) There was just not many Food Security event happened during the time when the live data was collected; (3) The labelers who reviewed Food Security articles did not perform as guided.   
Through reviewing samples with high food security category classification score we determined that there are missing labels due to improper annotator guidance and calibration.
This highlights the importance of performing regular data quality checks.

%% file: main.bbl
\begin{thebibliography}{15}
\providecommand{\natexlab}[1]{#1}

\bibitem[{Adauto et~al.(2023)Adauto, Jin, Sch{\"o}lkopf, Hope, Sachan, and Mihalcea}]{adauto-etal-2023-beyond}
Fernando Adauto, Zhijing Jin, Bernhard Sch{\"o}lkopf, Tom Hope, Mrinmaya Sachan, and Rada Mihalcea. 2023.
\newblock \href {https://doi.org/10.18653/v1/2023.findings-emnlp.31} {Beyond good intentions: Reporting the research landscape of {NLP} for social good}.
\newblock In \emph{Findings of the Association for Computational Linguistics: EMNLP 2023}, pages 415--438, Singapore. Association for Computational Linguistics.

\bibitem[{Conneau et~al.(2019)Conneau, Khandelwal, Goyal, Chaudhary, Wenzek, Guzm{\'a}n, Grave, Ott, Zettlemoyer, and Stoyanov}]{conneau2019unsupervised}
Alexis Conneau, Kartikay Khandelwal, Naman Goyal, Vishrav Chaudhary, Guillaume Wenzek, Francisco Guzm{\'a}n, Edouard Grave, Myle Ott, Luke Zettlemoyer, and Veselin Stoyanov. 2019.
\newblock Unsupervised cross-lingual representation learning at scale.
\newblock \emph{arXiv preprint arXiv:1911.02116}.

\bibitem[{Devlin et~al.(2018)Devlin, Chang, Lee, and Toutanova}]{devlin2018bert}
Jacob Devlin, Ming-Wei Chang, Kenton Lee, and Kristina Toutanova. 2018.
\newblock Bert: Pre-training of deep bidirectional transformers for language understanding.
\newblock \emph{arXiv preprint arXiv:1810.04805}.

\bibitem[{Duarte et~al.(2021)Duarte, Rawat, and Shah}]{duarte2021plm}
Kevin Duarte, Yogesh Rawat, and Mubarak Shah. 2021.
\newblock Plm: Partial label masking for imbalanced multi-label classification.
\newblock In \emph{Proceedings of the IEEE/CVF Conference on Computer Vision and Pattern Recognition}, pages 2739--2748.

\bibitem[{Elwell and Polikar(2011)}]{5975223}
Ryan Elwell and Robi Polikar. 2011.
\newblock \href {https://doi.org/10.1109/TNN.2011.2160459} {Incremental learning of concept drift in nonstationary environments}.
\newblock \emph{IEEE Transactions on Neural Networks}, 22(10):1517--1531.

\bibitem[{Jin et~al.(2021)Jin, Chauhan, Tse, Sachan, and Mihalcea}]{jin-etal-2021-good}
Zhijing Jin, Geeticka Chauhan, Brian Tse, Mrinmaya Sachan, and Rada Mihalcea. 2021.
\newblock \href {https://doi.org/10.18653/v1/2021.findings-acl.273} {How good is {NLP}? a sober look at {NLP} tasks through the lens of social impact}.
\newblock In \emph{Findings of the Association for Computational Linguistics: ACL-IJCNLP 2021}, pages 3099--3113, Online. Association for Computational Linguistics.

\bibitem[{Kshirsagar et~al.(2021)Kshirsagar, Robinson, Yang, Gholami, Klyuzhin, Mukherjee, Nasir, Ortiz, Oviedo, Tanner, Trivedi, Xu, Zhong, Dilkina, Dodhia, and Lavista~Ferres}]{10.1145/3461702.3462599}
Meghana Kshirsagar, Caleb Robinson, Siyu Yang, Shahrzad Gholami, Ivan Klyuzhin, Sumit Mukherjee, Md~Nasir, Anthony Ortiz, Felipe Oviedo, Darren Tanner, Anusua Trivedi, Yixi Xu, Ming Zhong, Bistra Dilkina, Rahul Dodhia, and Juan~M. Lavista~Ferres. 2021.
\newblock \href {https://doi.org/10.1145/3461702.3462599} {Becoming good at ai for good}.
\newblock In \emph{Proceedings of the 2021 AAAI/ACM Conference on AI, Ethics, and Society}, AIES '21, page 664–673, New York, NY, USA. Association for Computing Machinery.

\bibitem[{Lamba et~al.(2024)Lamba, Abilov, Zhang, Olson, Dambanemuya, B{\'a}rcia, Batista, Wille, Cahill, Tetreault, and Jaimes}]{lamba-etal-2024-humvi}
Hemank Lamba, Anton Abilov, Ke~Zhang, Elizabeth~M Olson, Henry~Kudzanai Dambanemuya, Jo{\~a}o~Cordovil B{\'a}rcia, David~S. Batista, Christina Wille, Aoife Cahill, Joel~R. Tetreault, and Alejandro Jaimes. 2024.
\newblock \href {https://doi.org/10.18653/v1/2024.findings-emnlp.743} {{H}um{VI}: A multilingual dataset for detecting violent incidents impacting humanitarian aid}.
\newblock In \emph{Findings of the Association for Computational Linguistics: EMNLP 2024}, pages 12705--12722, Miami, Florida, USA. Association for Computational Linguistics.

\bibitem[{Leetaru and Schrodt(2013)}]{leetaru2013gdelt}
Kalev Leetaru and Philip~A Schrodt. 2013.
\newblock Gdelt: Global data on events, location, and tone, 1979--2012.
\newblock In \emph{ISA annual convention}, volume~2, pages 1--49. Citeseer.

\bibitem[{Lisivick(2018)}]{lisivick2018newsapi}
Matt Lisivick. 2018.
\newblock Newsapi python library.

\bibitem[{Liu et~al.(2019)Liu, Ott, Goyal, Du, Joshi, Chen, Levy, Lewis, Zettlemoyer, and Stoyanov}]{liu2019roberta}
Yinhan Liu, Myle Ott, Naman Goyal, Jingfei Du, Mandar Joshi, Danqi Chen, Omer Levy, Mike Lewis, Luke Zettlemoyer, and Veselin Stoyanov. 2019.
\newblock Roberta: A robustly optimized bert pretraining approach.
\newblock \emph{arXiv preprint arXiv:1907.11692}.

\bibitem[{Sanh et~al.(2019)Sanh, Debut, Chaumond, and Wolf}]{sanh2019distilbert}
Victor Sanh, Lysandre Debut, Julien Chaumond, and Thomas Wolf. 2019.
\newblock Distilbert, a distilled version of bert: smaller, faster, cheaper and lighter.
\newblock \emph{arXiv preprint arXiv:1910.01108}.

\bibitem[{Shankar et~al.(2024)Shankar, Garcia, Hellerstein, and Parameswaran}]{Shankar_2024}
Shreya Shankar, Rolando Garcia, Joseph~M. Hellerstein, and Aditya~G. Parameswaran. 2024.
\newblock \href {https://doi.org/10.1145/3653697} {“we have no idea how models will behave in production until production”: How engineers operationalize machine learning}.
\newblock \emph{Proceedings of the ACM on Human-Computer Interaction}, 8(CSCW1):1–34.

\bibitem[{Toma{\v{s}}ev et~al.(2020)Toma{\v{s}}ev, Cornebise, Hutter, Mohamed, Picciariello, Connelly, Belgrave, Ezer, Haert, Mugisha et~al.}]{tomavsev2020ai}
Nenad Toma{\v{s}}ev, Julien Cornebise, Frank Hutter, Shakir Mohamed, Angela Picciariello, Bec Connelly, Danielle~CM Belgrave, Daphne Ezer, Fanny Cachat van~der Haert, Frank Mugisha, and 1 others. 2020.
\newblock Ai for social good: unlocking the opportunity for positive impact.
\newblock \emph{Nature Communications}, 11(1):2468.

\bibitem[{Zong et~al.(2021)Zong, Xia, Li, and Navigli}]{acl-2021-association}
Chengqing Zong, Fei Xia, Wenjie Li, and Roberto Navigli, editors. 2021.
\newblock \href {https://aclanthology.org/2021.acl-long.0/} {\emph{Proceedings of the 59th Annual Meeting of the Association for Computational Linguistics and the 11th International Joint Conference on Natural Language Processing (Volume 1: Long Papers)}}. Association for Computational Linguistics, Online.

\end{thebibliography}
